\documentclass[lettersize, journal]{IEEEtran}

\usepackage{color,array,amsthm}
\usepackage{graphicx}
\usepackage{kotex} 
\usepackage{multirow}
\usepackage{hyperref}
\usepackage{amsmath,amssymb,amsfonts}
\usepackage{orcidlink}

\begin{document}

\title{STAR: Scene- and Task-Aware 4D Radar Preprocessing Towards End-to-End Cognitive Radar}

\author{
Seung-Hyun Song\orcidlink{0009-0001-4813-779X},~\IEEEmembership{Student Member, IEEE},
Dong-Hee Paek\orcidlink{0000-0003-0008-3726},~\IEEEmembership{Member, IEEE},
and Seung-Hyun Kong\orcidlink{0000-0002-4753-1998},~\IEEEmembership{Senior Member, IEEE}
\thanks{
This work was supported by the Ministry of Trade, Industry and Resources (MOTIR) 
grant funded by the Korea government (No. RS-2025-25451359).
}
\thanks{
Seung-Hyun Song and Seung-Hyun Kong are with the Cho Chun Shik Graduate School of Mobility,
Korea Advanced Institute of Science and Technology (KAIST),
Daejeon 34141, Republic of Korea
(e-mail: \href{mailto:shyun@kaist.ac.kr}{shyun@kaist.ac.kr};
\href{mailto:skong@kaist.ac.kr}{skong@kaist.ac.kr}).
}
\thanks{
Dong-Hee Paek is with the Department of Future Mobility,
Korea University, Sejong 30019, Republic of Korea
(e-mail: \href{mailto:dhpaek@korea.ac.kr}{dhpaek@korea.ac.kr}).
}
\thanks{
Corresponding author: Seung-Hyun Kong.
}
}

\markboth{
Preprint submitted to arXiv
}{
Song \MakeLowercase{\textit{et al.}}: STAR: Scene- and Task-Aware 4D Radar Preprocessing
}

\maketitle

\begin{abstract}
Four-dimensional (4D) Radar has emerged as a key sensor for environmental perception, providing range, azimuth, elevation, and Doppler measurements while remaining robust to illumination changes and adverse weather conditions.
However, conventional Radar preprocessing methods, such as constant false alarm rate (CFAR) detection, select measurements primarily based on signal-level criteria and may therefore discard information valuable for downstream perception during point cloud generation. In addition, existing 4D Radar perception pipelines typically optimize Radar data processing and downstream perception independently, preventing task objectives from directly guiding the preprocessing stage. 
To address these limitations, we propose a Scene- and Task-Aware Radar (STAR) Preprocessor together with an end-to-end training framework. The STAR Preprocessor incorporates scene context and downstream task objectives to generate task-relevant Radar points, enabling the Radar representation to be optimized directly for perception. 
On the K-Radar benchmark, the proposed method achieves 74.3 \(AP_{\mathrm{BEV}}\), outperforming the previous state of the art by 5.6 $AP$ points. Furthermore, applying the task-relevant points generated by STAR to various existing 3D detectors improves detection performance in most evaluation settings and yields an overall positive average gain over point clouds produced by conventional preprocessing.
\end{abstract}

\begin{IEEEkeywords}4D Imaging Radar, Radar Preprocessing, 3D Object Detection, Cognitive Radar
\end{IEEEkeywords}

\section{INTRODUCTION}
Robust perception is essential for applications that require reliable situational awareness of the surrounding environment, including autonomous vehicles, intelligent robots, unmanned platforms, and military surveillance and reconnaissance systems~\cite{Intro1, Intro2, Intro3, Intro4, Intro5}. Cameras and LiDAR provide rich visual and geometric information; however, their performance can degrade under challenging environmental conditions, such as illumination changes, fog, rain, snow, and dust. In contrast, Radar uses electromagnetic waves and remains relatively robust to adverse weather and illumination variations~\cite{robust1, robust2, robust3, robust4}. In particular, recent four-dimensional (4D) Radar sensors provide high-resolution measurements in range, azimuth, elevation, and Doppler, expanding the role of conventional Radar from primarily range and velocity sensing to a key perception modality for 3D object detection, tracking, and scene understanding~\cite{4dradar1, 4dradar2, point3}.

\begin{figure}
    \centering    \includegraphics[width=1.0\linewidth]{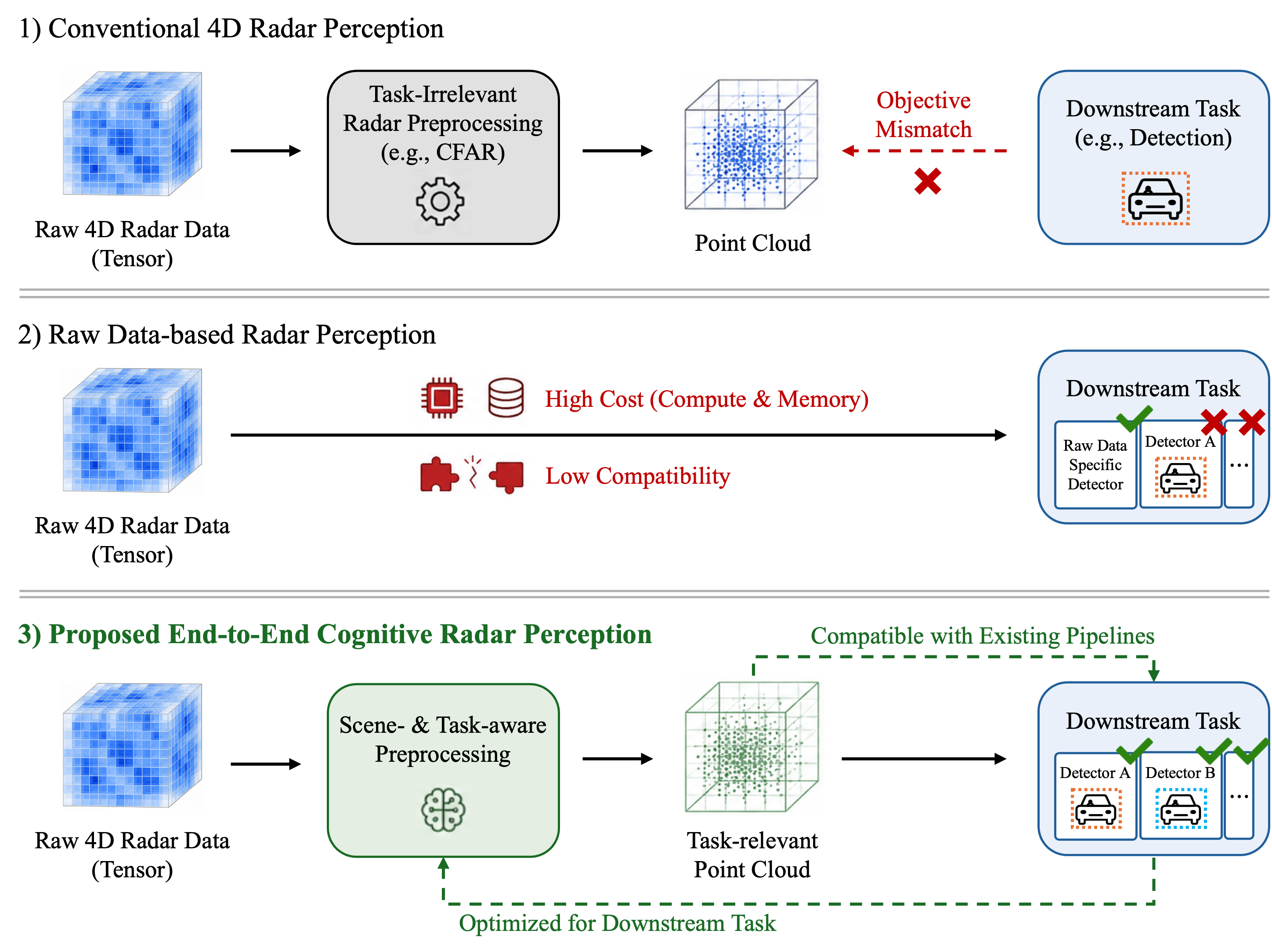}
    \caption{Comparison of 4D Radar perception paradigms. STAR enables scene- and task-aware preprocessing while preserving compatibility with existing point cloud detectors.}
    \label{fig1}
\end{figure}

However, current 4D Radar perception still relies heavily on conventional Radar preprocessing criteria that do not fully reflect the information requirements of downstream perception tasks. Traditional detection methods based on constant false alarm rate (CFAR) processing determine detection thresholds according to local clutter statistics, with the primary objective of suppressing false alarms while extracting reliable measurements that indicate the presence of targets~\cite{cfar1, cfar2, cfar3, cfar5}. Although such approaches are effective from a target detection perspective, they are not necessarily well suited for 3D perception, which requires preserving structural relationships among multiple Radar responses, such as object shape, boundaries, and spatial distributions. Indeed, previous studies have shown that simply varying the filtering level or data density of 4D Radar measurements can substantially affect downstream detection performance~\cite{enhancedkradar, noise, song, synthesis}. Therefore, preprocessing designed primarily around signal-level detectability can limit the extent to which the rich measurements provided by 4D Radar are exploited for high-level perception.

To mitigate the information loss and limited utilization of Radar measurements, existing 4D Radar perception studies have evolved along three main directions~\cite{related}. First, to extract more effective features from sparse Radar point clouds, various point-, voxel-, and pillar-based representations have been developed, often incorporating Radar-specific information such as Doppler velocity and Radar cross section (RCS)~\cite{pointpillars, rpfa, mfnet, radarpillarnet, mvfan, lxl, smurf, mufasa, mspfnet, dadan}. Second, to reduce information loss introduced during point cloud generation, several approaches directly process raw or higher-dimensional 4D Radar representations within perception networks rather than converting them into point clouds~\cite{rodnet, fft-radnet, adcnet, t-fftradnet, rade-net}. Third, other studies have focused on improving preprocessing strategies, including thresholding, point selection, and filtering, to more effectively preserve Radar measurements that are informative for downstream perception~\cite{kradar, rtnh+, rdn, corenet, hyperdet}.

Despite these advances, most existing 4D Radar perception pipelines still design Radar data processing and downstream perception independently. As illustrated Fig.~\ref{fig1}, the Radar representation is determined by predefined processing strategies, while the downstream task objective is not directly incorporated into the optimization of the Radar processing stage. This separation represents a key obstacle to extending cognitive Radar principles to high-level perception~\cite{cognitive}, where task objectives can guide Radar processing to optimize the utilization of sensing information. Moreover, because many existing Radar perception pipelines are designed around point cloud representations~\cite{point1, point2, point3}, methods that require raw or dense Radar tensors are not readily compatible with existing architectures, limiting their practical applicability.

In contrast to prior studies, this work addresses the fundamental structural limitation that Radar processing and downstream perception are optimized separately. To this end, we propose a Scene- and Task-Aware Radar (STAR) Preprocessor with an end-to-end framework that jointly optimizes Radar preprocessing and downstream perception by propagating detection objectives to the preprocessing stage.
The STAR Preprocessor learns spatial context from high-dimensional 4D Radar measurements and generates two complementary representations: a scene-aware bird's-eye-view (BEV) feature capturing object spatial distribution and foreground/background information, and task-relevant points selected for downstream object detection. Representing the selected measurements as sparse points reduces the computational and memory overhead of directly processing high-dimensional Radar tensors while maintaining compatibility with existing point cloud-based 3D detectors.
To enable the STAR Preprocessor to identify Radar information that is useful for downstream perception, we develop an end-to-end training framework that propagates the detection objective to the preprocessing stage. Multiple 3D detectors with different architectures are jointly used for supervision, reducing dependence on a specific detector and encouraging the learning of generally useful Radar representations.
Finally, we design a dedicated STAR Detector to exploit the scene-aware representation. It employs task-decoupled classification and localization with coarse-to-fine refinement for accurate 3D bounding box estimation.

On the K-Radar benchmark, which provides both 4D Radar tensors and preprocessed point clouds, STAR achieves 74.3 $AP_{\mathrm{BEV}}$, outperforming the previous state of the art by 5.6 $AP$ points. Furthermore, applying the task-relevant points generated by the STAR Preprocessor to various existing 3D detectors improves detection performance over conventionally preprocessed point clouds, demonstrating the effectiveness of optimizing Radar data processing itself according to scene context and downstream task objectives.

The main contributions of this work are summarized as follows:
\begin{itemize}
    \item We extend Radar preprocessing from conventional signal-level measurement selection to a task-aware information selection process for downstream perception, establishing a task-aware direction for 4D Radar data processing.
    \item We propose the STAR Preprocessor, which generates scene-aware features and task-relevant Radar points based on scene context and task objectives, together with the STAR Detector designed to effectively exploit these representations. The proposed framework achieves state-of-the-art performance on the K-Radar benchmark.
    \item We develop a joint end-to-end training framework that leverages multiple 3D detector objectives to reduce detector-specific bias in Radar preprocessing. The resulting task-relevant points improve performance across most detector–metric configurations and achieve substantial average gains over conventional point clouds.
\end{itemize}

The remainder of this paper is organized as follows. Section II reviews related work on 4D Radar perception and preprocessing. Section III presents the proposed STAR framework. Section IV provides experimental results and analyses on the K-Radar benchmark. Finally, Section V concludes the paper.

\section{Related Work}
\subsection{Point Cloud-Based 4D Radar 3D Object Detection}
Compared with LiDAR, 4D Radar point clouds are sparser, noisier, and more irregularly distributed, while providing Radar-specific attributes such as Doppler velocity and RCS. Existing studies therefore focus on improving point representations and feature encoding to better utilize the limited Radar points retained after preprocessing.

Early studies adopted PointPillars~\cite{pointpillars} to efficiently process Radar point clouds using pillar-based representations. RPFA-Net~\cite{rpfa} introduced attention to emphasize informative Radar pillar features and suppress noise, while RadarMFNet~\cite{mfnet} used ego-motion compensation and multi-frame accumulation to alleviate low point density. RadarPillarNet~\cite{radarpillarnet} transformed 4D Radar points into BEV pillar features for downstream detection. Subsequent methods further improved feature representation using Radar-specific attributes and structural context: MVFAN~\cite{mvfan} exploited Doppler, RCS, and multi-view features; SMURF~\cite{smurf} combined pillar features with spatial density representations; and MUFASA~\cite{mufasa} incorporated local geometry and multi-view context. More recently, MSPFNet~\cite{mspfnet} employed multi-scale pillar feature fusion, while DADAN~\cite{dadan} utilized velocity, intensity, and range-dependent point-density characteristics to reduce foreground-background ambiguity.

These studies have continuously improved 4D Radar object detection by extracting more effective features from pre-generated sparse Radar point clouds. However, the information available to the detector is inherently limited to measurements retained during preprocessing, making discarded measurements unavailable to downstream perception. Moreover, the selection of which Radar measurements to preserve is generally performed independently of the downstream detection objective.

\subsection{Raw and Tensor-Based 4D Radar 3D Object Detection}
Point cloud-based approaches are computationally efficient, but discard a substantial portion of Radar measurements during point cloud generation, limiting the use of information captured by the sensor. To address this issue, recent studies directly process low-level Radar data, such as raw ADC signals, Range-Doppler representations, and high-dimensional Radar tensors, within perception networks.

FFT-RadNet~\cite{fft-radnet} learned azimuth information from Range-Doppler representations, enabling object detection without explicitly constructing high-dimensional Radar tensors. ADCNet~\cite{adcnet} directly processed raw ADC measurements, allowing the network to learn information typically handled by conventional signal processing. T-FFTRadNet~\cite{t-fftradnet} further extended this direction with a transformer-based architecture for object detection from raw ADC signals. More recently, RADE-Net~\cite{rade-net} utilized 4D Radar tensors containing Range-Azimuth-Doppler-Elevation information and transformed them into efficient representations for 3D object detection.

However, ADC data and high-dimensional Radar tensors are substantially larger than sparse point clouds, resulting in higher memory and computational costs. They also typically require dedicated input encoders and detector architectures tailored to each low-level Radar representation.

\begin{figure*}
    \centering
    \includegraphics[width=1.0\linewidth]{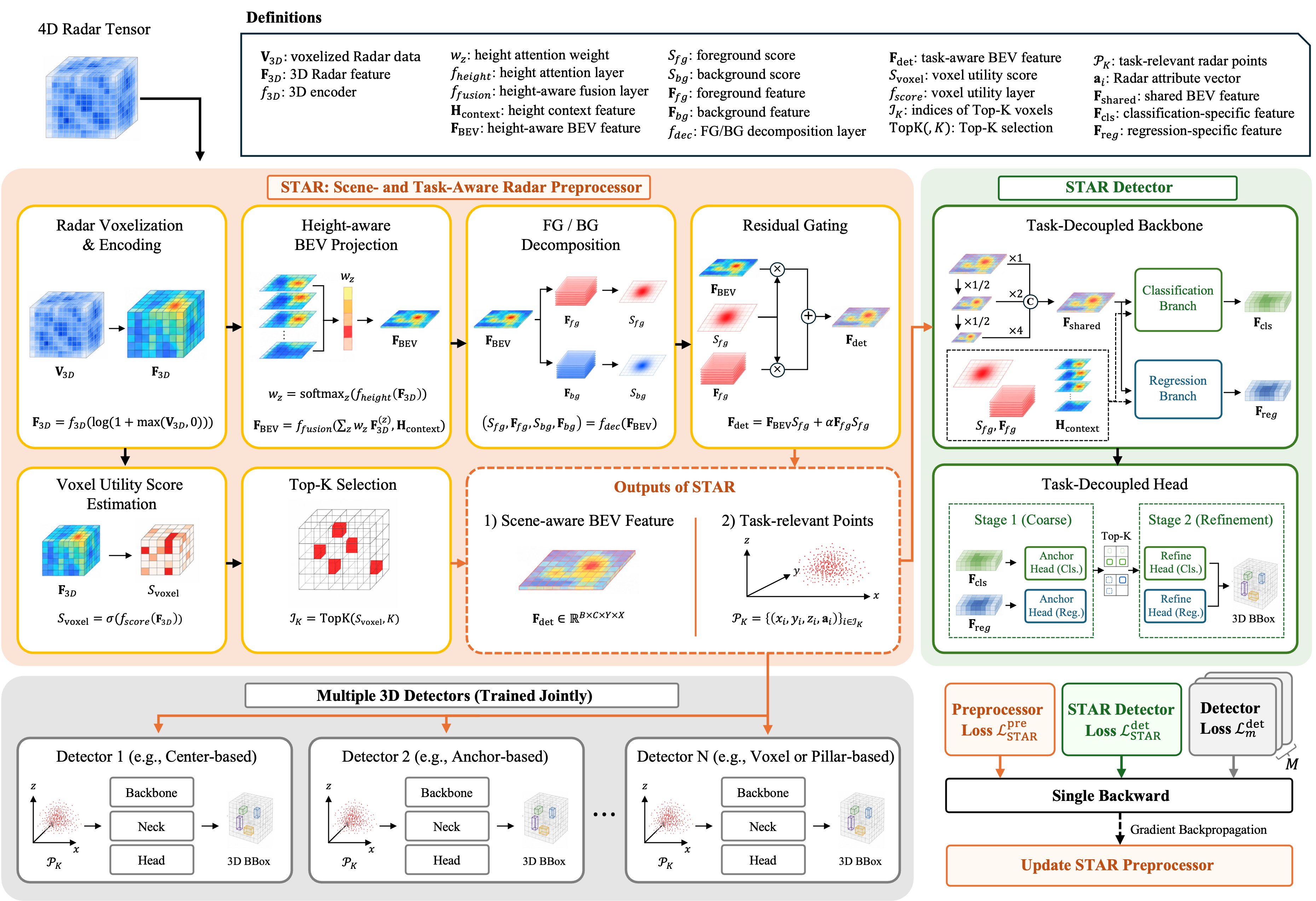}
    \caption{Overall architecture of the proposed STAR framework. The STAR Preprocessor generates scene-aware BEV features and task-relevant Radar points from 4D Radar tensors, while multi-detector supervision enables end-to-end optimization of the preprocessing strategy.}
    \label{fig2}
\end{figure*}

\subsection{Radar Preprocessing for 4D Radar Perception}
To preserve 4D Radar information in point cloud representations, several studies have focused on improving Radar preprocessing and point generation. RTNH~\cite{kradar} converted dense 4D Radar tensors into sparse representations, while Enhanced K-Radar~\cite{enhancedkradar} analyzed the relationship between Radar density and detection performance to derive an efficient sparse representation. RTNH+~\cite{rtnh+} further employed two-level preprocessing with different filtering levels to retain richer Radar measurements. More recently, learning-based preprocessing methods have also been explored. RDN~\cite{rdn} denoised and complemented noisy Radar point clouds, CORENet~\cite{corenet} learned to extract informative Radar points using LiDAR supervision, and HyperDet~\cite{hyperdet} generated enhanced point representations through Radar point aggregation and refinement for use with existing detectors.

Although these studies demonstrate the importance of Radar preprocessing, most rely on predefined filtering strategies or post-process already generated point clouds. This motivates a preprocessing approach that directly selects and preserves task-relevant information from high-dimensional Radar measurements according to downstream objectives, while maintaining compatibility with existing point cloud-based detectors.

\section{Proposed Method}
\subsection{Overview}
Fig.~\ref{fig2} illustrates the overall architecture of the proposed framework. We integrate Radar preprocessing and downstream perception into a unified training pipeline so that the downstream detection objective can directly guide the preprocessing stage. The input 4D Radar tensor is transformed by the STAR Preprocessor into representations that reflect both scene context and task requirements.

The STAR Preprocessor generates two complementary representations. First, it produces a scene-aware BEV feature that captures the spatial structure and scene context of the 4D Radar tensor, which is used as contextual information by the STAR Detector. Second, it estimates the task relevance of each Radar measurement and selects task-relevant Radar points for downstream detection. Since these points retain a sparse point cloud format, they can be directly used by existing point-, pillar-, and voxel-based 3D detectors.

To reduce dependence of the STAR Preprocessor on a specific detector architecture, we employ multiple 3D detectors with different architectures for joint supervision within the end-to-end training framework. Detection objectives from each detector are propagated to the shared STAR Preprocessor, encouraging it to learn Radar representations that are broadly useful across different detectors. The following subsections describe the end-to-end learning framework, STAR Preprocessor, STAR Detector, and training objectives in sequence.

\subsection{End-to-End Framework for Scene- and Task-Aware Radar Preprocessing}
\label{sec:3.2}
In the proposed framework, the scene-aware representation and task-relevant Radar points produced by the STAR Preprocessor are fed to the STAR Detector and multiple 3D detectors with different architectures. The detection objectives from these detectors are backpropagated to the shared STAR Preprocessor, allowing the importance of each Radar measurement to be learned based on its contribution to downstream object detection rather than solely on signal-level criteria.

When the preprocessing stage is trained with a single detector, the resulting Radar representation may become overly dependent on architecture-specific characteristics such as voxelization, feature encoding, receptive fields, and detection heads. To mitigate this issue, we employ $M$ 3D detectors with heterogeneous architectures as task supervision. By jointly propagating gradients from multiple detectors to the shared STAR Preprocessor, the model is encouraged to preserve Radar information that is broadly useful across detectors rather than measurements favored by a specific architecture. These detectors are used only as training supervision to reduce detector-specific dependency in preprocessing, not as an inference ensemble for combining predictions.

Meanwhile, the hard Top-$K$ selection used for task-relevant point extraction is non-differentiable, preventing the detection objectives of the auxiliary detectors from being directly propagated to the voxel utility scores through the selected-point pathway. To enable end-to-end optimization, we introduce a differentiable surrogate gating pathway during training. Specifically, the 3D voxel utility scores are projected onto the BEV feature space of each auxiliary detector and converted into continuous gating weights that modulate the intermediate detector features. The resulting detection gradients are backpropagated through the gating weights to the voxel utility scores, allowing the STAR Preprocessor to learn which Radar measurements are useful for downstream detection without differentiating through the hard Top-$K$ operation. As the utility scores are progressively optimized, the measurements selected by Top-$K$ also change accordingly throughout training. Further details on the voxel utility estimation and task-relevant point extraction are provided in Section~III-\ref{sec:3.3}.

Since the task-relevant Radar points retain the conventional point cloud format, existing detectors can be trained and evaluated without modification. The joint training objective is described in Section~III-\ref{sec:3.5}.

\subsection{STAR Preprocessor}
\label{sec:3.3}
The STAR Preprocessor takes a dense Cartesian Radar power volume constructed from 4D Radar measurements as input and generates a scene-aware BEV feature for the STAR Detector and a task-relevant Radar point cloud compatible with existing 3D detectors. To this end, the input Radar measurements are transformed into Cartesian voxel space and sequentially processed through 3D scene encoding, height-aware projection, foreground/background decomposition, residual feature gating, and voxel-level relevance estimation.

Since Radar power exhibits a large dynamic range, logarithmic compression is first applied to reduce the dominance of strong reflections and preserve weak responses. A 3D encoder then extracts spatially contextualized features from the compressed Radar volume.
\begin{equation}
\mathbf{F}_{3D}
= f_{3D} \left(
\log\left(1+\max(\mathbf{V}_{3D},0)\right)
\right),
\end{equation}
where $\mathbf{V}_{3D}$ denotes the Radar power volume constructed in Cartesian space, and $\mathbf{F}_{3D}$ represents the 3D feature incorporating the spatial context of surrounding Radar responses. Unlike conventional voxel-wise thresholding, each voxel feature in $\mathbf{F}_{3D}$ aggregates information from neighboring horizontal and vertical Radar responses, enabling a scene-aware representation beyond individual reflections.

The 3D feature is then projected onto the BEV space using a height-aware projection. Specifically, the height attention layer $f_{\mathrm{height}}$ estimates the importance of each height from $\mathbf{F}_{3D}$ at every spatial location, and generates the attention weights $w_z$ through a softmax operation along the height dimension:
\begin{equation}
w_z(x,y)
= \operatorname{softmax}_z \left(
f_{\mathrm{height}}(\mathbf{F}_{3D})
\right),
\end{equation}
where $f_{\mathrm{height}}$ is a $1 \times 1 \times 1$ convolutional layer that converts each voxel feature into a height-importance logit.

To compensate for vertical structure that may be lost during height-weighted projection, STAR additionally constructs four height descriptors with complementary characteristics. $\mathbf{F}_{\max}$ preserves the strongest feature response along the height dimension, while $\mathbf{F}_{\mathrm{top}}$ aggregates dominant height features with high attention weights. In addition, $\mathbf{H}_{\mathrm{disp}}$ represents how concentrated or dispersed the height attention is across different heights, and $\mathbf{H}_{\mathrm{peak}}$ indicates the height location of the dominant response. These descriptors are defined as follows.
\begin{equation}
\begin{aligned}
\mathbf{F}_{\max}
&=
\max_{z}\mathbf{F}_{3D}^{(z)}, \\
\mathbf{F}_{\mathrm{top}}
&=
\sum_{z\in\operatorname{TopK}_{z}(w)}
\bar{w}_{z}\mathbf{F}_{3D}^{(z)},\\
\mathbf{H}_{\mathrm{disp}}
&=
-\frac{1}{\log Z}
\sum_{z} w_z \log w_z,\\
\mathbf{H}_{\mathrm{peak}}
&=
\frac{\operatorname*{arg\,max}_{z} w_z}{Z-1},
\end{aligned}
\end{equation}
where $\bar{w}_{z}$ denotes the attention weight renormalized over the selected Top-$K$ heights, and $Z$ is the number of voxels along the height dimension of the 3D Radar feature. $\mathbf{F}_{\max}$ and $\mathbf{F}_{\mathrm{top}}$ preserve dominant vertical feature responses, while $\mathbf{H}_{\mathrm{disp}}$ and $\mathbf{H}_{\mathrm{peak}}$ provide vertical dispersion and dominant height information, respectively. These complementary height descriptors are concatenated to form the height context representation $\mathbf{H}_{\mathrm{context}}$.
\begin{equation}
\mathbf{H}_{\mathrm{context}}
=
\operatorname{Concat}
\left(
\mathbf{F}_{\max},
\mathbf{F}_{\mathrm{top}},
\mathbf{H}_{\mathrm{disp}},
\mathbf{H}_{\mathrm{peak}}
\right).
\end{equation}
Finally, the height-weighted 3D feature and height context are combined to generate the height-aware BEV feature $\mathbf{F}_{\mathrm{BEV}}$.
\begin{equation}
\mathbf{F}_{\mathrm{BEV}}
=
f_{\mathrm{fusion}}
\left(
\operatorname{Concat}
\left[
\sum_{z} w_z \mathbf{F}_{3D}^{(z)},
\mathbf{H}_{\mathrm{context}}
\right]
\right).
\end{equation}

\begin{figure}
    \centering
    \includegraphics[width=1.0 \linewidth]{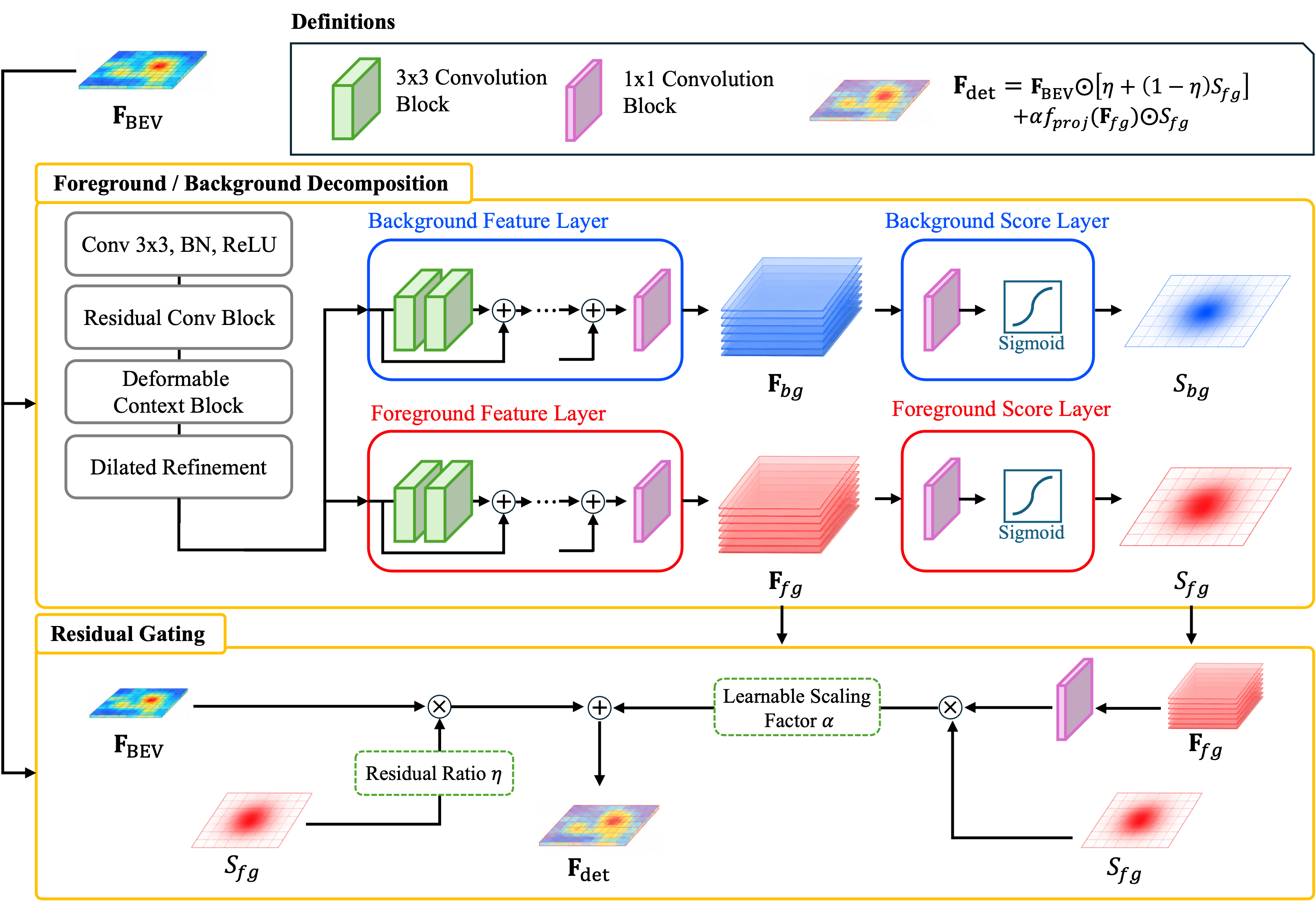}
    \caption{Detailed architecture of the foreground/background decomposition and residual gating modules in the STAR Preprocessor.}
    \label{fig3}
\end{figure}

The generated $\mathbf{F}_{\mathrm{BEV}}$ contains both object-related responses and background responses from the surrounding scene. Rather than simply suppressing the background, STAR explicitly separates the two components using a foreground/background decomposition module $f_{\mathrm{dec}}$. As shown in Fig.~\ref{fig3}, the module first extracts a shared contextual feature from $\mathbf{F}_{\mathrm{BEV}}$, and then generates foreground and background features and scores through two independent branches.

\begin{equation}
\left(
\mathbf{S}_{\mathrm{fg}},
\mathbf{F}_{\mathrm{fg}},
\mathbf{S}_{\mathrm{bg}},
\mathbf{F}_{\mathrm{bg}}
\right)
=
f_{\mathrm{dec}}
\left(
\mathbf{F}_{\mathrm{BEV}}
\right),
\end{equation}
where $\mathbf{F}_{\mathrm{fg}}$ and $\mathbf{F}_{\mathrm{bg}}$ denote foreground- and background-specific contextual features, respectively, while $\mathbf{S}_{\mathrm{fg}}$ and $\mathbf{S}_{\mathrm{bg}}$ are score maps representing the foreground and background relevance at each spatial location. Each branch consists of a feature layer and a score layer to independently model object-related responses and scene-dependent background responses.

The feature passed to the STAR Detector is then generated through residual gating using foreground information. As illustrated at the bottom of Fig.~\ref{fig3}, the residual gating consists of a base gating path that reweights the original BEV feature according to the foreground score and a context injection path that selectively enhances it with the foreground contextual feature.
\begin{equation}
\mathbf{F}_{\mathrm{det}}
=
\mathbf{F}_{\mathrm{BEV}}
\odot
\left[
\eta+(1-\eta)\mathbf{S}_{\mathrm{fg}}
\right]
+
\alpha f_{\mathrm{proj}}(\mathbf{F}_{\mathrm{fg}})
\odot\mathbf{S}_{\mathrm{fg}},
\end{equation}
where $\eta$ is a residual ratio that preserves a minimum amount of feature information in low-score regions, $\alpha$ is a learnable scaling factor that controls the contribution of foreground context, and $f_{\mathrm{proj}}$ aligns the channel dimension of $\mathbf{F}_{\mathrm{fg}}$ with $\mathbf{F}_{\mathrm{BEV}}$. The first term preserves the original scene feature without completely suppressing low-score regions, while the second selectively enhances foreground-related contextual information. As a result, $\mathbf{F}_{\mathrm{det}}$ retains global scene information while emphasizing foreground cues useful for object detection.

Meanwhile, STAR estimates a voxel utility score from $\mathbf{F}_{3D}$ to represent the downstream task relevance of each Radar voxel.
\begin{equation}
\mathbf{S}_{\mathrm{voxel}}
=
\sigma\left(
f_{\mathrm{score}}(\mathbf{F}_{3D})
\right).
\end{equation}
Unlike $\mathbf{S}_{\mathrm{fg}}$, which is used to construct the scene-aware BEV feature, $\mathbf{S}_{\mathrm{voxel}}$ represents the task relevance of each Radar measurement for deciding which measurements are retained in the final point cloud. Through the end-to-end supervision described in Section~III-\ref{sec:3.2}, the utility scores are optimized according to the downstream detection objective. The $K$ voxels with the highest utility scores are then selected to extract the task-relevant Radar point cloud.
\begin{equation}
\mathcal{J}_{K}
=
\operatorname{TopK}
\left(
\mathbf{S}_{\mathrm{voxel}}, K
\right).
\end{equation}
The selected voxels are converted into metric coordinates and represented in a Radar point format compatible with existing detectors.
\begin{equation}
\mathcal{P}_{K}
=
\left\{
(x_i,y_i,z_i,\mathbf{a}_i)
\mid i \in \mathcal{J}_{K}
\right\},
\end{equation}
where $\mathbf{a}_i$ denotes the Radar attribute vector associated with the selected voxel, which may include Radar power, Doppler velocity, and other available measurement attributes. In this work, we use $\mathbf{a}_i = \rho_i$, where $\rho_i$ denotes the Radar power of the corresponding voxel.

Consequently, the STAR Preprocessor simultaneously produces the task-aware BEV feature $\mathbf{F}_{\mathrm{det}}$ for the STAR Detector and the task-relevant point cloud $\mathcal{P}_{K}$ for existing point cloud-based detectors.

\subsection{STAR Detector}
The STAR Detector takes $\mathbf{F}_{\mathrm{det}}$ generated by the STAR Preprocessor as input and predicts 3D bounding boxes.

First, $\mathbf{F}_{\mathrm{det}}$ is processed by a multi-scale backbone to extract BEV features with different receptive fields. The features from each scale are aligned to the same spatial resolution and fused through concatenation, producing a shared feature $\mathbf{F}_{\mathrm{shared}}$ that captures both local object information and global scene context. The shared feature is then separated into classification and regression branches to generate task-specific features.

\begin{equation}
\mathbf{F}_{\mathrm{cls}}
=
f_{\mathrm{cls}}
\left(
\mathbf{F}_{\mathrm{shared}}
\right),
\end{equation}

\begin{equation}
\mathbf{F}_{\mathrm{reg}}
=
f_{\mathrm{reg}}
\left(
\operatorname{Concat}
\left[
\mathbf{F}_{\mathrm{shared}},
\mathbf{C}_{\mathrm{STAR}}
\right]
\right),
\end{equation}
where $\mathbf{C}_{\mathrm{STAR}}$ denotes contextual information extracted by the STAR Preprocessor, including foreground scores and features as well as height-aware context. This information is additionally provided to the regression branch to enhance scene-level cues for estimating object location and size. The classification branch extracts discriminative features $\mathbf{F}_{\mathrm{cls}}$ for distinguishing objects from background, while the regression branch generates $\mathbf{F}_{\mathrm{reg}}$ using foreground distribution and height information for box localization.

The STAR Detector then employs a coarse-to-fine two-stage detection head. Since the sparse and irregular spatial responses of Radar measurements can hinder accurate box localization, Stage 1 generates dense 3D box proposals from the task-decoupled features and selects high-confidence candidates. Stage 2 further refines the confidence scores and box parameters of the selected proposals, reducing localization errors caused by Radar spatial uncertainty while preserving broad detection coverage.

\subsection{Training Objectives}
\label{sec:3.5}
The STAR framework jointly optimizes the objectives of the STAR Preprocessor and downstream detectors to learn both scene-aware representations and task-relevant Radar point selection. First, the STAR Preprocessor loss for learning foreground/background decomposition is defined as follows.
\begin{equation}
\begin{aligned}
\mathcal{L}_{\mathrm{STAR}}^{\mathrm{pre}}
={}&
\lambda_{\mathrm{fg\text{-}obj}}
\mathcal{L}_{\mathrm{fg\text{-}obj}}
+
\lambda_{\mathrm{fg\text{-}sm}}
\mathcal{L}_{\mathrm{fg\text{-}sm}}
+
\lambda_{\mathrm{bg}}
\mathcal{L}_{\mathrm{bg}}
\\
&+
\lambda_{\mathrm{bg\text{-}sm}}
\mathcal{L}_{\mathrm{bg\text{-}sm}}
+
\lambda_{\mathrm{sep}}
\mathcal{L}_{\mathrm{sep}}
+
\lambda_{\mathrm{rec}}
\mathcal{L}_{\mathrm{rec}},
\end{aligned}
\end{equation}
where the foreground target $\mathbf{G}_{\mathrm{fg}}$ is constructed by combining Gaussian responses centered at the center and front/rear locations of each oriented GT 3D box, while the background target $\mathbf{G}_{\mathrm{bg}}$ is defined as the normalized BEV Radar power excluding the dilated foreground regions.
$\mathcal{L}_{\mathrm{fg\text{-}obj}}$ is a focal-weighted binary cross-entropy loss for learning foreground objectness between $\mathbf{S}_{\mathrm{fg}}$ and $\mathbf{G}_{\mathrm{fg}}$, while $\mathcal{L}_{\mathrm{fg\text{-}sm}}$ and $\mathcal{L}_{\mathrm{bg\text{-}sm}}$ are Total Variation losses for enforcing spatial consistency of the foreground and background scores, respectively. $\mathcal{L}_{\mathrm{bg}}$ is a Smooth L1 loss that encourages $\mathbf{S}_{\mathrm{bg}}$ to reconstruct $\mathbf{G}_{\mathrm{bg}}$, whereas $\mathcal{L}_{\mathrm{sep}}$ and $\mathcal{L}_{\mathrm{rec}}$ suppress overlap between the foreground and background scores and preserve the overall Radar scene response, respectively. In our experiments, $\lambda_{\mathrm{fg\text{-}obj}}$, $\lambda_{\mathrm{fg\text{-}sm}}$, $\lambda_{\mathrm{bg}}$, $\lambda_{\mathrm{bg\text{-}sm}}$, $\lambda_{\mathrm{sep}}$, and $\lambda_{\mathrm{rec}}$ are set to $1.0$, $0.05$, $0.2$, $0.02$, $0.05$, and $0.1$, respectively.

The STAR Detector loss is defined by jointly considering coarse detection and proposal refinement as follows.
\begin{equation}
\begin{aligned}
\mathcal{L}_{\mathrm{STAR}}^{\mathrm{det}}
={}&
\lambda_{\mathrm{cls}}
\mathcal{L}_{\mathrm{cls}}
+
\lambda_{\mathrm{box}}
\mathcal{L}_{\mathrm{box}}
+
\lambda_{\mathrm{dir}}
\mathcal{L}_{\mathrm{dir}}
\\
&+
\lambda_{\mathrm{ref}}
\mathcal{L}_{\mathrm{ref}}
+
\lambda_{\mathrm{obj}}
\mathcal{L}_{\mathrm{obj}},
\end{aligned}
\end{equation}
where $\mathcal{L}_{\mathrm{cls}}$, $\mathcal{L}_{\mathrm{box}}$, and $\mathcal{L}_{\mathrm{dir}}$ denote the classification, 3D bounding box regression, and direction estimation losses of the coarse stage, respectively. $\mathcal{L}_{\mathrm{ref}}$ and $\mathcal{L}_{\mathrm{obj}}$ denote the box refinement and objectness losses for the selected proposals.

\begin{table*}[t]
\rmfamily
\centering
\caption{Comparison with Radar-based state-of-the-art methods on the K-Radar dataset for the Sedan class at an IoU threshold of 0.3.}
\label{tab:sota_comparison}
\begin{tabular}{c|c|cccccccc}
\hline\hline
Network & Metric & Total & Normal & Overcast & Fog & Rain & Sleet & Light Snow & Heavy Snow \\ 
\hline\hline

\multirow{2}{*}{RTNH\cite{kradar}}
& $AP_{\mathrm{3D}}$  & 47.4 & 49.9 & 56.7 & 52.8 & 42.0 & 41.5 & 50.6 & 44.5 \\
& $AP_{\mathrm{BEV}}$ & 58.4 & 58.5 & 64.2 & 76.2 & 58.4 & 60.3 & 57.6 & 56.6 \\
\hline

\multirow{2}{*}{RTNH+\cite{rtnh+}}
& $AP_{\mathrm{3D}}$  & 57.6 & 59.88 & 70.61 & 73.03 & 48.25 & 48.87 & 72.08 & 58.65 \\
& $AP_{\mathrm{BEV}}$ & 65.7 & 64.25 & 77.36 & 84.35 & 59.47 & 56.75 & 73.78 & 65.98 \\
\hline

\multirow{2}{*}{RADE-Net\cite{rade-net}}
& $AP_{\mathrm{3D}}$  & 64.1 & 60.7 & 72.0 & 85.4 & 55.4 & \textbf{63.6$^{*}$} & 68.6 & \textbf{67.6$^{*}$} \\
& $AP_{\mathrm{BEV}}$ & 68.7 & 65.2 & 74.6 & \textbf{91.3$^{**}$} & \textbf{63.1$^{**}$} & \textbf{67.9$^{**}$} & 74.1 & \textbf{68.7$^{**}$} \\
\hline

\multirow{2}{*}{STAR (Ours)}
& $AP_{\mathrm{3D}}$
& \textbf{\shortstack{64.8$^{*}$(+0.7)}}
& \textbf{63.9$^{*}$} & 68.4 & \textbf{87.1$^{*}$}
& \textbf{57.0$^{*}$} & 54.2 & \textbf{78.9$^{*}$} & 59.4 \\

& $AP_{\mathrm{BEV}}$
& \textbf{\shortstack{74.3$^{**}$(+5.6)}}
& \textbf{73.4$^{**}$} & \textbf{79.9$^{**}$} & 89.7
& 61.0 & 60.5 & \textbf{87.5$^{**}$} & 67.0 \\
\hline\hline
\multicolumn{10}{l}{\footnotesize
$^{*}$ Best performance in $AP_{\mathrm{3D}}$;
$^{**}$ Best performance in $AP_{\mathrm{BEV}}$.}
\end{tabular}
\end{table*}

Finally, to reduce the dependence of the preprocessing stage on a specific detector, we jointly optimize the task objectives from the STAR Detector and $M$ heterogeneous auxiliary detectors. Let $\mathcal{L}_{m}^{\mathrm{det}}$ denote the detection objective of the $m$th auxiliary detector. The overall loss is defined as follows.
\begin{equation}
\mathcal{L}_{\mathrm{total}}
=
\frac{
\lambda_{\mathrm{STAR}}
\left(
\mathcal{L}_{\mathrm{STAR}}^{\mathrm{det}}
+
\mathcal{L}_{\mathrm{STAR}}^{\mathrm{pre}}
\right)
+
\sum_{m=1}^{M}
\lambda_m
\mathcal{L}_{m}^{\mathrm{det}}
}{
\lambda_{\mathrm{STAR}}
+
\sum_{m=1}^{M}\lambda_m
},
\end{equation}
where $\lambda_{\mathrm{STAR}}$ and $\lambda_m$ control the task contributions of the STAR Detector and the $m$th auxiliary detector, respectively. We set $\lambda_{\mathrm{STAR}}=1.0$ and $\lambda_m=0.25$ for all auxiliary detectors. All objectives are jointly optimized in a single backward pass to update the STAR Preprocessor and each detector branch. No explicit voxel-level ground-truth supervision is applied to $\mathbf{S}_{\mathrm{voxel}}$; instead, its task relevance is learned solely from the downstream detection objectives.

\section{Experiments}
\subsection{Dataset}
We evaluate the proposed STAR framework on the K-Radar dataset~\cite{kradar}, which provides dense 4D Radar tensors. Unlike conventional methods that directly use preprocessed point clouds, STAR aims to learn task-relevant Radar representations from high-dimensional Radar measurements. Therefore, K-Radar is well suited to our study because it provides access to dense Radar representations before point cloud generation.

K-Radar provides 34,994 frames of dense 4D Radar tensors collected under diverse road and weather conditions, with each tensor consisting of 64 Doppler, 256 range, 107 azimuth, and 37 elevation bins. Instead of directly processing the polar-coordinate 4D tensor, we use the Cartesian Radar power cube provided by K-Radar. The cube is constructed by averaging Radar power along the Doppler dimension and interpolating the resulting range--azimuth--elevation tensor onto a Cartesian $Z$--$Y$--$X$ grid. The original Cartesian cube has a spatial resolution of $0.4$~m and a size of $150 \times 400 \times 250$. We restrict the region of interest to $x \in [0,72)$~m, $y \in [-6.4,6.4)$~m, and $z \in [-2,7.6)$~m, resulting in a single-channel Radar power tensor of size $24 \times 32 \times 180$ as the input to the STAR Preprocessor.

Although K-Radar provides Doppler information, velocity ambiguity may occur due to its limited unambiguous Doppler range, and the Cartesian power cube used in our experiments is constructed by averaging along the Doppler dimension. Accordingly, we use $\mathbf{a}_i=\rho_i$ for the Radar attribute vector, excluding Doppler velocity from the STAR point features.

\subsection{Experiment Setup}
\textbf{Implementation Details.}  
All models were implemented using PyTorch 1.12.1, CUDA 11.3, and cuDNN 8.3.2, and all experiments were conducted on a single NVIDIA GeForce RTX 3090 GPU with 24~GB of memory. The proposed model was trained using the AdamW optimizer with an initial learning rate of $1\times10^{-3}$, a minimum learning rate of $1\times10^{-4}$, and a weight decay of $0.01$. The momentum parameters were set to $\beta_1=0.9$ and $\beta_2=0.999$, and a cosine annealing scheduler was used to gradually decay the learning rate. Both standalone training of the STAR Detector and joint training with multiple detectors were performed for 30 epochs, with batch sizes of 8 and 4, respectively.
For joint training, six heterogeneous 3D detectors ($M=6$), namely RTNH~\cite{kradar}, RPFA~\cite{rpfa}, RadarPillarNet~\cite{radarpillarnet}, MVFAN~\cite{mvfan}, SMURF~\cite{smurf}, and DADAN~\cite{dadan}, were used as auxiliary supervision networks.

\textbf{Evaluation Metrics.}
We evaluate 3D object detection performance on the Sedan and Bus or Truck classes of the K-Radar benchmark. We use bird's-eye-view average precision, $AP_{\mathrm{BEV}}$, and 3D average precision, $AP_{\mathrm{3D}}$. $AP_{\mathrm{BEV}}$ is computed using the oriented IoU between rotated BEV bounding boxes, while $AP_{\mathrm{3D}}$ is based on the volumetric IoU between rotated 3D bounding boxes. Both metrics are evaluated at IoU thresholds of $0.3$ and $0.5$ following the KITTI-style evaluation protocol.

\begin{table*}[]
\rmfamily
\centering
\caption{Comparison of 3D object detection performance using the K-Radar original point cloud and the proposed STAR point cloud across multiple detectors. All detectors were retrained using the same dataset split and training protocol, with only the input point cloud replaced.}
\label{tab2}
\begin{tabular}{c|c|cccc|cccc}
\hline \hline
\multirow{3}{*}{Network} & \multirow{3}{*}{Class} & \multicolumn{4}{c|}{K-Radar Original Point Cloud} & \multicolumn{4}{c}{\textbf{STAR Point Cloud (Ours)}} \\ \cline{3-10} 
 &  & \multicolumn{2}{c|}{IoU 0.3} & \multicolumn{2}{c|}{IoU 0.5} & \multicolumn{2}{c|}{IoU 0.3} & \multicolumn{2}{c}{IoU 0.5} \\ \cline{3-10} 
 &  & \multicolumn{1}{c|}{$AP_{\mathrm{BEV}}$} & \multicolumn{1}{c|}{$AP_{\mathrm{3D}}$} & $AP_{\mathrm{BEV}}$ & $AP_{\mathrm{3D}}$ & $AP_{\mathrm{BEV}}$ & \multicolumn{1}{c|}{$AP_{\mathrm{3D}}$} & $AP_{\mathrm{BEV}}$ & $AP_{\mathrm{3D}}$ \\ \hline \hline
\multirow{2}{*}{RTNH\cite{kradar}} & Sedan & 58.4 & \multicolumn{1}{c|}{47.4} & 43.18 & 15.60 & \textbf{\begin{tabular}[c]{@{}c@{}}66.27\\ (+7.87)\end{tabular}} & \multicolumn{1}{c|}{\textbf{\begin{tabular}[c]{@{}c@{}}62.85\\ (+15.45)\end{tabular}}} & \textbf{\begin{tabular}[c]{@{}c@{}}54.54\\ (+11.36)\end{tabular}} & \textbf{\begin{tabular}[c]{@{}c@{}}33.75\\ (+18.15)\end{tabular}} \\
 & Bus or Truck & 45.3 & \multicolumn{1}{c|}{34.4} & - & - & \textbf{\begin{tabular}[c]{@{}c@{}}47.93\\ (+2.63)\end{tabular}} & \multicolumn{1}{c|}{\textbf{\begin{tabular}[c]{@{}c@{}}34.80\\ (+0.40)\end{tabular}}} & 32.31 & 16.78 \\ \hline
\multirow{2}{*}{RPFA$^{\dagger}$\cite{rpfa}} & Sedan & 54.00 & \multicolumn{1}{c|}{45.16} & 43.43 & 23.17 & \textbf{\begin{tabular}[c]{@{}c@{}}62.89\\ (+8.89)\end{tabular}} & \multicolumn{1}{c|}{\textbf{\begin{tabular}[c]{@{}c@{}}53.82\\ (+8.66)\end{tabular}}} & \textbf{\begin{tabular}[c]{@{}c@{}}50.85\\ (+7.42)\end{tabular}} & \textbf{\begin{tabular}[c]{@{}c@{}}29.69\\ (+6.52)\end{tabular}} \\
 & Bus or Truck & 31.28 & \multicolumn{1}{c|}{26.70} & 19.48 & 7.20 & \textbf{\begin{tabular}[c]{@{}c@{}}40.86\\ (+9.58)\end{tabular}} & \multicolumn{1}{c|}{\textbf{\begin{tabular}[c]{@{}c@{}}38.40\\ (+11.70)\end{tabular}}} & \textbf{\begin{tabular}[c]{@{}c@{}}35.02\\ (+15.54)\end{tabular}} & \textbf{\begin{tabular}[c]{@{}c@{}}19.28\\ (+12.08)\end{tabular}} \\ \hline
\multirow{2}{*}{RadarPillarNet$^{\dagger}$\cite{radarpillarnet}} & Sedan & \textbf{58.19} & \multicolumn{1}{c|}{55.61} & \textbf{47.55} & \textbf{32.38} & 58.13 & \multicolumn{1}{c|}{\textbf{\begin{tabular}[c]{@{}c@{}}55.74\\ (+0.13)\end{tabular}}} & 46.98 & 31.69 \\
 & Bus or Truck & 48.82 & \multicolumn{1}{c|}{41.11} & \textbf{35.31} & \textbf{21.57} & \textbf{\begin{tabular}[c]{@{}c@{}}49.14\\ (+0.32)\end{tabular}} & \multicolumn{1}{c|}{\textbf{\begin{tabular}[c]{@{}c@{}}41.88\\ (+0.77)\end{tabular}}} & 33.69 & 17.13 \\ \hline
\multirow{2}{*}{MVFAN$^{\dagger}$\cite{mvfan}} & Sedan & 63.62 & \multicolumn{1}{c|}{53.52} & 45.90 & 25.24 & \textbf{\begin{tabular}[c]{@{}c@{}}65.52\\ (+1.90)\end{tabular}} & \multicolumn{1}{c|}{\textbf{\begin{tabular}[c]{@{}c@{}}56.67\\ (+3.15)\end{tabular}}} & \textbf{\begin{tabular}[c]{@{}c@{}}53.19\\ (+7.29)\end{tabular}} & \textbf{\begin{tabular}[c]{@{}c@{}}31.54\\ (+6.30)\end{tabular}} \\
 & Bus or Truck & 38.20 & \multicolumn{1}{c|}{26.01} & 24.56 & 8.95 & \textbf{\begin{tabular}[c]{@{}c@{}}47.58\\ (+9.38)\end{tabular}} & \multicolumn{1}{c|}{\textbf{\begin{tabular}[c]{@{}c@{}}35.80\\ (+9.79)\end{tabular}}} & \textbf{\begin{tabular}[c]{@{}c@{}}32.46\\ (+7.90)\end{tabular}} & \textbf{\begin{tabular}[c]{@{}c@{}}17.31\\ (+8.36)\end{tabular}} \\ \hline
\multirow{2}{*}{SMURF$^{\dagger}$\cite{smurf}} & Sedan & 56.78 & \multicolumn{1}{c|}{48.13} & 46.16 & 25.31 & \textbf{\begin{tabular}[c]{@{}c@{}}58.04\\ (+1.26)\end{tabular}} & \multicolumn{1}{c|}{\textbf{\begin{tabular}[c]{@{}c@{}}55.49\\ (+7.36)\end{tabular}}} & \textbf{\begin{tabular}[c]{@{}c@{}}46.94\\ (+0.78)\end{tabular}} & \textbf{\begin{tabular}[c]{@{}c@{}}31.66\\ (+6.35)\end{tabular}} \\
 & Bus or Truck & 47.12 & \multicolumn{1}{c|}{34.58} & \textbf{32.48} & \textbf{17.44} & \textbf{\begin{tabular}[c]{@{}c@{}}48.35\\ (+1.23)\end{tabular}} & \multicolumn{1}{c|}{\textbf{\begin{tabular}[c]{@{}c@{}}36.25\\ (+1.67)\end{tabular}}} & 30.21 & 12.67 \\ \hline
\multirow{2}{*}{DADAN$^{\dagger}$\cite{dadan}} & Sedan & 57.51 & \multicolumn{1}{c|}{53.10} & 45.50 & 25.65 & \textbf{\begin{tabular}[c]{@{}c@{}}63.86\\ (+6.35)\end{tabular}} & \multicolumn{1}{c|}{\textbf{\begin{tabular}[c]{@{}c@{}}54.48\\ (+1.38)\end{tabular}}} & \textbf{\begin{tabular}[c]{@{}c@{}}51.85\\ (+6.35)\end{tabular}} & \textbf{\begin{tabular}[c]{@{}c@{}}29.39\\ (+3.74)\end{tabular}} \\
 & Bus or Truck & 44.45 & \multicolumn{1}{c|}{\textbf{34.07}} & 28.27 & 12.34 & \textbf{\begin{tabular}[c]{@{}c@{}}45.96\\ (+1.51)\end{tabular}} & \multicolumn{1}{c|}{32.92} & \textbf{\begin{tabular}[c]{@{}c@{}}30.50\\ (+2.23)\end{tabular}} & \textbf{\begin{tabular}[c]{@{}c@{}}16.51\\ (+4.17)\end{tabular}} \\ \hline \hline
\multicolumn{10}{l}{\footnotesize
$^{\dagger}$ indicates reproduced results.}
\end{tabular}
\end{table*}

\subsection{Comparison with State-of-the-Art Methods}
Table~\ref{tab:sota_comparison} compares the proposed STAR with existing 4D Radar tensor-based 3D object detection methods on the K-Radar benchmark. All results are evaluated at an IoU threshold of $0.3$ using $AP_{\mathrm{3D}}$ and $AP_{\mathrm{BEV}}$ as performance metrics. STAR achieves the best overall performance with $64.8$ $AP_{\mathrm{3D}}$ and $74.3$ $AP_{\mathrm{BEV}}$. Compared with the previous best-performing RADE-Net, STAR improves $AP_{\mathrm{3D}}$ by $0.7$ AP points and $AP_{\mathrm{BEV}}$ by $5.6$ AP points. These results demonstrate the effectiveness of selectively preserving task-relevant information from high-dimensional Radar measurements while jointly exploiting scene-aware representations.

Across different weather conditions, STAR also shows strong performance under Normal, Overcast, and Light Snow conditions, achieving $78.9$ $AP_{\mathrm{3D}}$ and $87.5$ $AP_{\mathrm{BEV}}$ in Light Snow. Although RADE-Net~\cite{rade-net} performs better under some adverse weather conditions, STAR achieves the highest total $AP_{\mathrm{3D}}$ and $AP_{\mathrm{BEV}}$ across all conditions. This indicates that the proposed scene- and task-aware Radar preprocessing improves overall Radar perception performance under diverse driving and weather conditions.

\subsection{Effectiveness of Task-Relevant Radar Points}
Table~\ref{tab2} evaluates the effectiveness of the task-relevant Radar points generated by the STAR Preprocessor by comparing the original K-Radar point cloud and the STAR point cloud using the same 3D detectors. The STAR point cloud is constructed by selecting the top 8,192 measurements according to the voxel utility scores, while retaining a sparse representation compatible with existing detectors. After joint training, the STAR point clouds are generated in advance, and each detector is independently retrained using either the original or STAR point cloud under the same training protocol. We evaluate RTNH~\cite{kradar}, RPFA~\cite{rpfa}, RadarPillarNet~\cite{radarpillarnet}, MVFAN~\cite{mvfan}, SMURF~\cite{smurf}, and DADAN~\cite{dadan} using $AP_{\mathrm{BEV}}$ and $AP_{\mathrm{3D}}$ for the Sedan and Bus or Truck classes at IoU thresholds of $0.3$ and $0.5$.
Across all comparable evaluation settings, the STAR point cloud improves performance by an average of approximately $4.8$ AP points. In particular, at an IoU threshold of $0.5$, $AP_{\mathrm{BEV}}$ and $AP_{\mathrm{3D}}$ improve by approximately $4.95$ and $5.07$ AP points on average, respectively, indicating that STAR preserves Radar measurements that are more informative for accurate object localization.

Although performance decreases in a few detector--metric settings, the STAR point cloud outperforms the original point cloud across most evaluation configurations. These improvements, achieved by replacing only the input point representation without modifying the detector architecture, demonstrate that STAR can serve as a task-aware preprocessing front-end for existing point cloud-based 4D Radar perception systems.

\begin{table*}[]
\rmfamily
\centering
\caption{Comparison of data size and computational efficiency between the K-Radar original point cloud and the proposed STAR point cloud. Latency and GPU memory usage are measured across multiple 3D detectors, and the reduction rate indicates the relative decrease achieved by STAR.}
\label{tab3}
\begin{tabular}{cc|ccc}
\hline \hline
\multicolumn{2}{c|}{} & \begin{tabular}[c]{@{}c@{}}K-Radar Original\\ Point Cloud\end{tabular} & \textbf{\begin{tabular}[c]{@{}c@{}}STAR Point \\ Cloud (Ours)\end{tabular}} & \begin{tabular}[c]{@{}c@{}}Reduction Rate\\ (\%)\end{tabular} \\ \hline \hline
\multicolumn{2}{c|}{Points/Frame} & 18,162$\pm$8,783 & \textbf{8,192} & 54.9 \\ \hline 
\multicolumn{2}{c|}{Data/1,000 Frames (MB)} & 290.7 & \textbf{131.2} & 54.9 \\ \hline \hline
\multicolumn{1}{c|}{\multirow{2}{*}{RTNH~\cite{kradar}}} & Latency (ms) & 12.08 & \textbf{11.19} & 7.37 \\
\multicolumn{1}{c|}{} & GPU Memory (MB) & 200.91 & \textbf{185.62} & 7.61 \\ \hline
\multicolumn{1}{c|}{\multirow{2}{*}{RPFA~\cite{rpfa}}} & Latency (ms) & 9.98 & \textbf{9.52} & 4.61 \\
\multicolumn{1}{c|}{} & GPU Memory (MB) & 118.54 & \textbf{66.81} & 43.64 \\ \hline
\multicolumn{1}{c|}{\multirow{2}{*}{RadarPillarNet~\cite{radarpillarnet}}} & Latency (ms) & 9.70 & \textbf{8.75} & 9.79 \\
\multicolumn{1}{c|}{} & GPU Memory (MB) & 196.10 & \textbf{138.98} & 29.13 \\ \hline
\multicolumn{1}{c|}{\multirow{2}{*}{MVFAN~\cite{mvfan}}} & Latency (ms) & 25.56 & \textbf{24.10} & 5.72 \\
\multicolumn{1}{c|}{} & GPU Memory (MB) & 766.00 & \textbf{760.45} & 0.72 \\ \hline
\multicolumn{1}{c|}{\multirow{2}{*}{SMURF~\cite{smurf}}} & Latency (ms) & 124.04 & \textbf{40.06} & 67.70 \\
\multicolumn{1}{c|}{} & GPU Memory (MB) & 282.08 & \textbf{104.07} & 63.11 \\ \hline
\multicolumn{1}{c|}{\multirow{2}{*}{DADAN~\cite{dadan}}} & Latency (ms) & 23.64 & \textbf{19.42} & 17.85 \\
\multicolumn{1}{c|}{} & GPU Memory (MB) & 2570.18 & \textbf{428.31} & 83.34 \\ \hline \hline
\end{tabular}
\end{table*}

Table~\ref{tab3} further compares the computational efficiency of downstream detectors using the STAR point cloud. The original K-Radar point cloud contains an average of $18{,}162 \pm 8{,}783$ points per frame, whereas STAR selects only $8{,}192$ points based on task relevance, reducing both the number of points and data volume by approximately $54.9\%$. The fixed point budget also provides a more predictable computational workload regardless of scene-dependent variations in point density.

For this evaluation, the STAR point clouds are generated and stored in advance, and the downstream detectors are subsequently trained and evaluated using these preprocessed point clouds without running the STAR Preprocessor. Therefore, the reported latency and GPU memory usage measure the computational cost of the downstream detectors only. Using the STAR point cloud reduces both latency and GPU memory usage across all detectors. In particular, the latency of SMURF decreases by $67.70\%$, from $124.04$~ms to $40.06$~ms, while the GPU memory usage of DADAN decreases by $83.34\%$, from $2570.18$~MB to $428.31$~MB. These results demonstrate that the compact task-relevant point representation reduces unnecessary point processing and improves the computational efficiency of existing point cloud-based detectors without modifying their architectures.

Overall, the results in Table~\ref{tab2} and Table~\ref{tab3} demonstrate that STAR simultaneously improves detection performance and computational efficiency in existing point cloud-based 4D Radar perception pipelines. These findings support the effectiveness of task-aware preprocessing, which learns to select Radar information according to scene context and downstream task requirements rather than relying on predefined signal-level criteria.

\subsection{Qualitative Analysis}
\begin{figure*}
    \centering
    \includegraphics[width=1.0\linewidth]{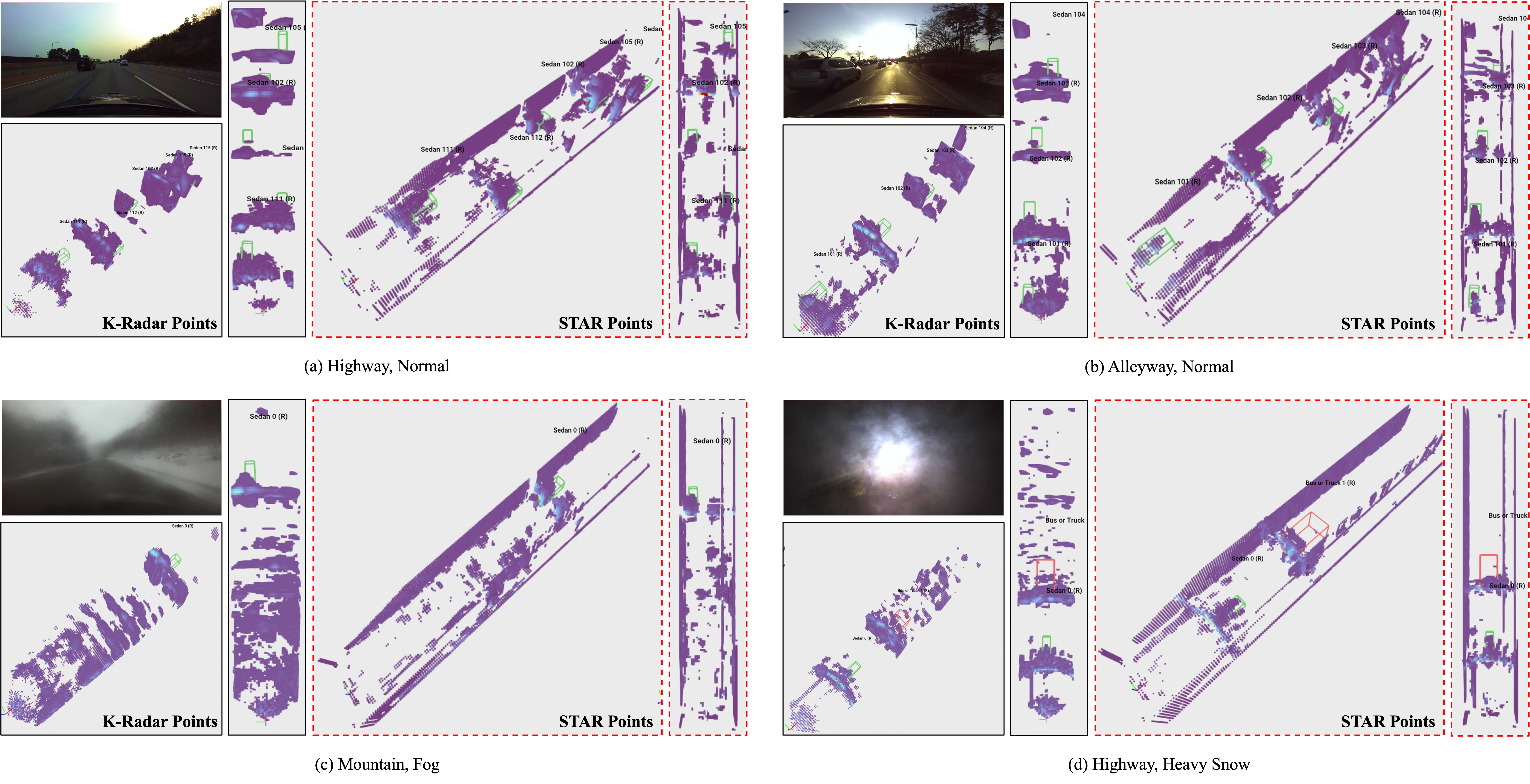}
    \caption{Qualitative comparison between the K-Radar original point cloud and the proposed STAR point cloud under diverse driving and weather conditions: (a) highway in normal weather, (b) alleyway in normal weather, (c) mountain road in fog, and (d) highway in heavy snow.}
    \label{fig4}
\end{figure*}

Fig.~\ref{fig4} qualitatively compares the original K-Radar point cloud with the task-relevant point cloud generated by the STAR Preprocessor under various driving and weather conditions. In Highway and Alleyway scenes under Normal conditions, STAR preserves Radar measurements around annotated objects, resulting in a clearer representation of their spatial structure. In particular, it suppresses high-power non-object responses concentrated near the Radar sensor while retaining informative measurements around objects. Similar behavior is observed under adverse weather conditions such as Fog and Heavy Snow, indicating that STAR selectively preserves object-related Radar information while suppressing noise and background responses.

STAR selects a fixed number of $K=8{,}192$ voxels per frame to maintain a consistent computational budget. Therefore, when the number of high-utility object-related measurements is insufficient, lower-utility background voxels may be included in the remaining point budget, occasionally producing wall-like point distributions near the RoI boundary. These points are not newly generated Radar responses, but residual background measurements selected by the fixed Top-$K$ operation. The effects of the point budget and such boundary points on detection performance are further analyzed in Section~IV-\ref{sec:4.6}.

\begin{table}[]
\centering
\caption{Component-wise ablation study of the proposed STAR framework on the K-Radar dataset.}
\label{tab4}
\resizebox{1.0\columnwidth}{!}{%
\begin{tabular}{ccc|c|cc}
\hline\hline
\multicolumn{3}{c|}{STAR Preprocessor} & \multirow{2}{*}{\begin{tabular}[c]{@{}c@{}}STAR\\ Detector\end{tabular}} & \multirow{2}{*}{$AP_{\mathrm{BEV}}$} & \multirow{2}{*}{$AP_{\mathrm{3D}}$} \\ \cline{1-3}
\begin{tabular}[c]{@{}c@{}}Height-aware\\ Projection\end{tabular} & \begin{tabular}[c]{@{}c@{}}FG/BG\\ Decomposition\end{tabular} & \begin{tabular}[c]{@{}c@{}}Residual\\ Gating\end{tabular} &  &  &  \\ \hline \hline
 & \checkmark & \checkmark& \checkmark& 72.9 & 62.6 \\
\checkmark&  & \checkmark& \checkmark& 67.7 & 64.7 \\
\checkmark& \checkmark& & \checkmark& 68.1 & 64.7 \\
\checkmark& \checkmark& \checkmark&  & 67.8 & 64.6 \\ \hline
\checkmark& \checkmark& \checkmark& \checkmark& \textbf{74.3} & \textbf{64.8} \\ \hline \hline
\end{tabular}
}
\end{table}

\subsection{Ablation Study}
\label{sec:4.6}
\textbf{Point Budget Analysis.}
To determine an appropriate point budget for the STAR point cloud, we conduct an ablation study with $K={4{,}096, 8{,}192, 16{,}384}$. For the Sedan class at an IoU threshold of $0.3$, $K=8{,}192$ achieves $74.30$ $AP_{\mathrm{BEV}}$ and $64.82$ $AP_{\mathrm{3D}}$. With $K=4{,}096$, $AP_{\mathrm{3D}}$ remains comparable at $64.88$, while $AP_{\mathrm{BEV}}$ decreases to $68.52$, indicating that a limited point budget may not sufficiently preserve the spatial extent of objects. In contrast, increasing $K$ to $16{,}384$ reduces $AP_{\mathrm{BEV}}$ and $AP_{\mathrm{3D}}$ to $68.23$ and $64.69$, respectively, suggesting that an excessive point budget may retain background measurements with relatively low task relevance. Accordingly, we set $K=8{,}192$ as the final point budget to balance task-relevant information preservation and background suppression.

\textbf{Proposed STAR Pipeline Analysis.}
To analyze the contribution of each STAR component, we evaluate variants without Height-aware Projection, Foreground/Background Decomposition, Residual Gating, and the STAR Detector. As shown in Table~\ref{tab4}, the full STAR model achieves the best overall performance with $74.3$ $AP_{\mathrm{BEV}}$ and $64.8$ $AP_{\mathrm{3D}}$. Removing Height-aware Projection causes the largest drop in $AP_{\mathrm{3D}}$ to $62.6$, while removing Foreground/Background Decomposition and Residual Gating reduces $AP_{\mathrm{BEV}}$ to $67.7$ and $68.1$, respectively. When the STAR Detector is excluded, the task-aware BEV feature is evaluated using a standard BEV backbone followed by an anchor-based detection head, resulting in $67.8$ $AP_{\mathrm{BEV}}$. These results indicate that the proposed components work complementarily to improve the scene-aware representation and overall detection performance.

\textbf{Boundary Filtering Analysis.}
To analyze the effect of the wall-like point distributions observed near the RoI boundary, we compare the original STAR point cloud with filtered STAR point cloud obtained by removing the boundary points. Specifically, boundary points are defined as points lying within a one-voxel-wide region along the outer boundary of the RoI. As shown in Table~\ref{tab5}, most detectors maintain similar $AP_{\mathrm{BEV}}$ and $AP_{\mathrm{3D}}$ before and after filtering, while some detectors show slight improvements.

These results suggest that, because the STAR Preprocessor is trained using downstream detection objectives, measurements important for object detection are assigned high task relevance, whereas the wall-like boundary points correspond to residual measurements with relatively low task utility. Therefore, these boundary points are not a major factor in overall detection performance, and the performance gains of STAR mainly arise from effectively preserving task-relevant Radar information around objects.

\begin{table}[]
\centering
\caption{Ablation study on the effect of boundary filtering for STAR point clouds across multiple 3D detectors.}
\label{tab5}
\begin{tabular}{c|cc|cc}
\hline \hline
\multirow{2}{*}{Network} & \multicolumn{2}{c|}{\begin{tabular}[c]{@{}c@{}}Original\\ STAR Point Cloud\end{tabular}} & \multicolumn{2}{c}{\begin{tabular}[c]{@{}c@{}}Filtered\\ STAR Point Cloud\end{tabular}} \\ \cline{2-5} 
 & AP\_BEV & AP\_3D & AP\_BEV & AP\_3D \\ \hline \hline
RTNH~\cite{kradar} & \textbf{66.27} & \textbf{62.85} & 65.89 & 62.57 \\ \hline
RPFA~\cite{rpfa} & \textbf{62.89} & 53.82 & 57.45 & \textbf{55.01} \\ \hline
RadarPillarNet~\cite{radarpillarnet} & 58.13 & 55.74 & \textbf{58.55} & \textbf{56.06} \\ \hline
MVFAN~\cite{mvfan} & \textbf{65.52} & \textbf{56.67} & 65.10 & 56.31 \\ \hline
SMURF~\cite{smurf} & \textbf{58.04} & \textbf{55.49} & 57.65 & 55.14 \\ \hline
DADAN~\cite{dadan} & 63.86 & 54.48 & \textbf{64.44} & \textbf{55.27} \\ \hline \hline
\end{tabular}
\end{table}

\section{Conclusion}
In this work, we proposed the STAR framework to bridge the gap between conventional 4D Radar preprocessing and downstream perception by learning Radar representations according to scene context and task requirements. STAR propagates downstream detection objectives to the preprocessing stage to generate scene-aware features and task-relevant Radar points. On the K-Radar benchmark, STAR achieves $74.3$ $AP_{\mathrm{BEV}}$ and $64.8$ $AP_{\mathrm{3D}}$, improving the previous best performance by $5.6$ and $0.7$ AP points, respectively. Applying the STAR point cloud to various existing 3D detectors improves detection performance across most evaluation settings while reducing the computational cost of downstream point cloud processing, demonstrating the effectiveness and applicability of the proposed preprocessing approach. Overall, this work advances task-aware 4D Radar processing by optimizing the Radar data representation according to scene context and downstream task objectives, moving beyond conventional signal-driven preprocessing. This approach may further serve as a foundation for Cognitive Radar systems that more tightly integrate perception objectives with sensing and signal-processing stages.

\bibliographystyle{IEEEtran}
\bibliography{ref}

\begin{IEEEbiography}[{\includegraphics[width=1in,height=1.25in,clip,keepaspectratio]{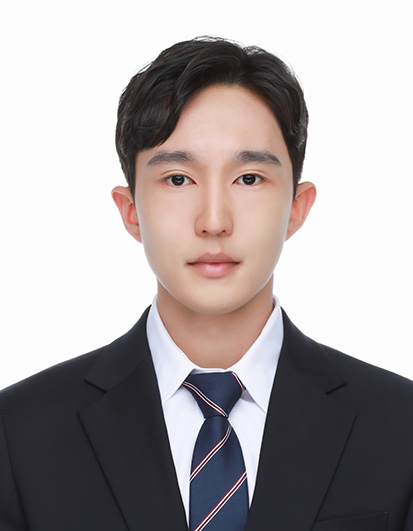}}]{Seung-Hyun Song}
received the B.S. degree in Automobile and IT Convergence from Kookmin University, Seoul, South Korea, in 2024, and the M.S. degree from the Graduate School of Advanced Security Science and Technology, Korea Advanced Institute of Science and Technology (KAIST), Daejeon, South Korea, in 2026. He is currently pursuing the Ph.D. degree with the CCS Graduate School of Mobility, KAIST. His research interests include 4D Radar, AI-based perception, and autonomous vehicles.
\end{IEEEbiography}

\begin{IEEEbiography}
[{\includegraphics[width=1in,height=1.25in,clip,keepaspectratio]{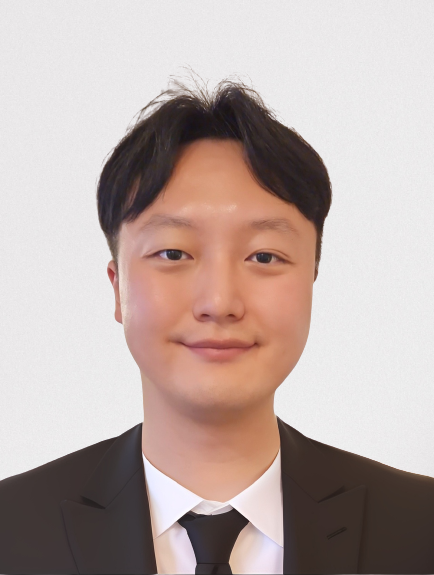}}]{Dong-Hee Paek} is currently an Assistant Professor with the Department of Future Mobility, Korea University, Sejong, Korea. He received the B.S. degree in robotics and intelligent systems from Kwangwoon University, Seoul, Korea, in 2019, and the M.S. degree in robotics and the Ph.D. degree in mobility and robotics from the Korea Advanced Institute of Science and Technology (KAIST), Daejeon, Korea, in 2021 and 2025, respectively. From 2024 to 2025, he was with Zeta Mobility, Daejeon, Korea, as a Senior Research Engineer, where he conducted research and development on 4D Radar perception and sensor fusion AI while pursuing the Ph.D. degree. From 2025 to 2026, he was a Postdoctoral Research Fellow with KAIST. His current research interests include 4D Radar signal processing and perception, multi-modal sensor fusion, and AI-based robotic perception.
\end{IEEEbiography}

\begin{IEEEbiography}[{\includegraphics[width=1in,height=1.25in,clip,keepaspectratio]{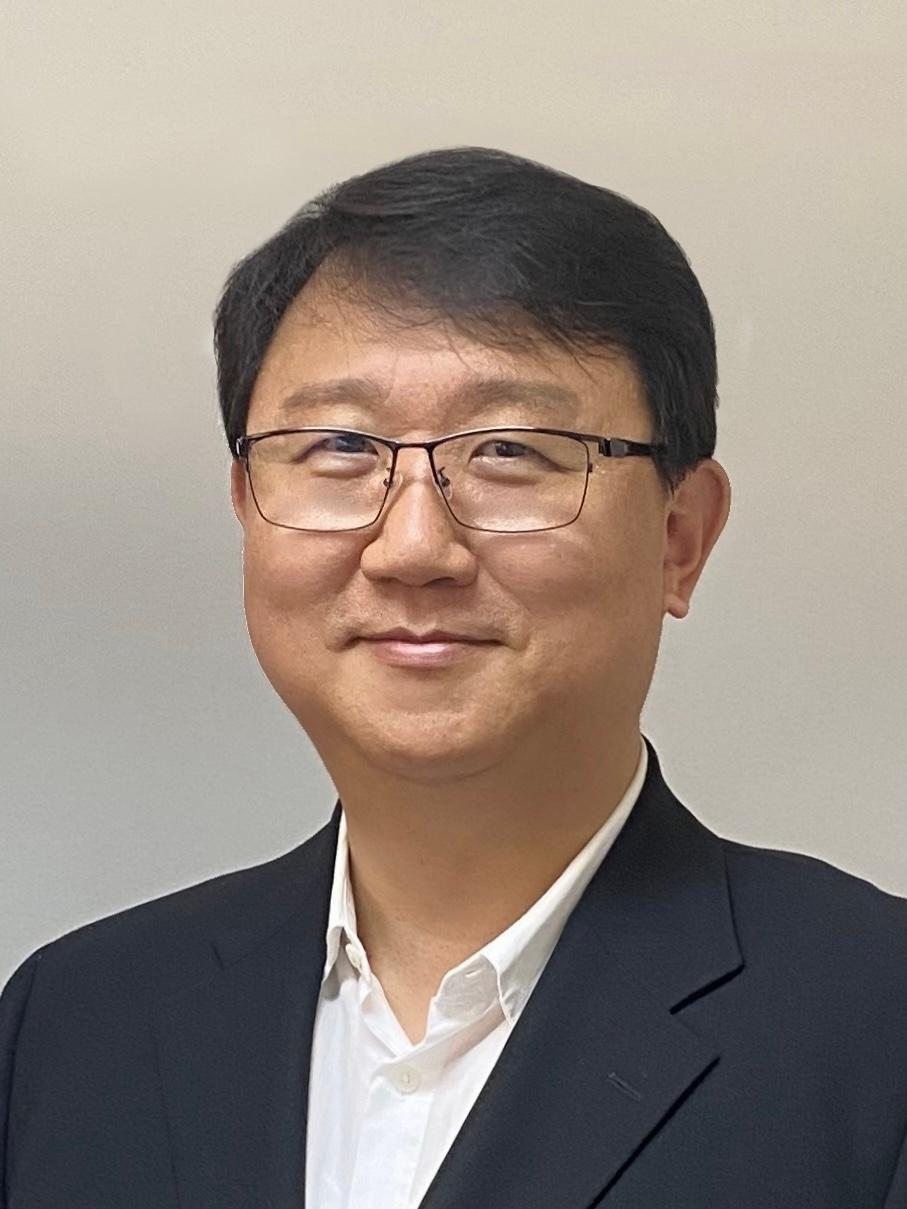}}]{Seung-Hyun Kong}
(M'06–SM'16) is an Associate Professor in the CCS Graduate School of Mobility, Korea Advanced Institute of Science and Technology (KAIST), where he has been a faculty member since 2010. He received a B.S. in Electronics Engineering from Sogang University, Seoul, Korea, in 1992; an M.S. in Electrical and Computer Engineering from Polytechnic University (now part of NYU), New York, in 1994; and a Ph.D. in Aeronautics and Astronautics from Stanford University in 2005. From 1997 to 2004 and again from 2006 to 2010, he worked at Samsung Electronics (Telecommunication Research Center), Nexpilot (Korea), Polaris Wireless (San Jose), and Qualcomm Corporate R\&D (San Diego), focusing on advanced R\&D in mobile communication, wireless positioning, and assisted GNSS. His current research interests include deep neural networks for 4D Radar-based perception and localization, sensor fusion, reinforcement learning for end-to-end autonomous driving, and compression for embedded systems. He has authored over 100 journal and conference papers, holds 14 patents, and his group received the President's Award at the 2018 International Student Autonomous Driving Competition in Korea. He was Program Co-Chair of IEEE ITSC 2019 (New Zealand), General Chair of IEEE IV 2024 (Korea), and has served as an Associate Editor of IEEE T-ITS since 2017.
\end{IEEEbiography}

\end{document}